# Manipulator Control of the Robotized TMS System with Incurved TMS Coil Case

Jaewoo Kim, Gi-Hun Yang*

*Abstract*—*Objective*: This study shows the force/torque control strategy for the robotized TMS system whose TMS coil's floor is incurved. The strategy considered the adhesion and friction between the coil and the subject's head. *Methods*: Hybrid position/force control and proportional torque were used for the strategy. The force magnitude applied for the force control was scheduled by the error between the coil's current position and the target point. *Results*: The larger desired force for the force controller makes the error quickly. By scheduling the force magnitude applied for the force control, the low error between the coil's current and target positions is maintained with the relatively small force after the larger force is applied for around 10 seconds. The proportional torque made the adhesion better by locating the contact are between the coil and the head close to the coil. I was shown by checking the $\tau_c/F_c$ value from the experimental results. While the head slowly moved away from the coil during the TMS treatment, the coil still interacted with the head. Using that characteristic, the coil could locate the new target point using the force/torque strategy without any trajectory planning. *Conclusion*: The proposed force/torque controller enhanced the adhesion between the incurved TMS coil and the subject's head. It also reduced the error quickly by scheduling the magnitude of the force applied. *Significance*: This study proposes the robotized TMS system's force/torque control strategy considering the physical characteristics from the contact between the incurved TMS coil case and the subject's head.

*Index Terms*—Hybrid position/force control, Robot manipulation, Torque control, Transcranial Magnetic Stimulation

## I. INTRODUCTION

TRANSCRANIAL Magnetic Stimulation (TMS) is a non-invasive and safe technique that makes the induced current in the brain by passing the brief and high-intensity magnetic field generated by the magnetic coil [1-2], [9]. This technique is used in many areas, for example, brain mapping [3-5], [12] learning brain-behavior relations [6-8], and medical therapeutic applications [9-11].

The professional operator usually manually manipulates the TMS coil using TMS systems. However, adjusting the TMS coil manually has a limit for reducing the error between the coil position and the target stimulation point. Moreover, because each TMS treatment takes tens of times [5], [13], [14], it is hard to compensate for the subject's head movement by holding the coil by hand continuously. Additional devices can be used [15], [16] to solve that problem. However, because those devices make subjects uncomfortable from their bad wearing sensation, they cause stress. Because this stress makes subjects secrete cortisol, which affects neuronal activity [18], it can cause variability in the efficacy of the TMS [17], [19].

To supplement those shortages, the robotic TMS system in which a robot manipulator controls the TMS coil is proposed [17], [20]-[25]. Because robot manipulators have high positioning accuracy and repeatability, they can make better coil positioning accuracy. Furthermore, because neuronavigation in the robotic TMS system can estimate the subject's head's position and target stimulation site, the operator does not have to continuously adjust the coil for a long time.

Some studies have used the force control strategy of the robot manipulator to improve the robotized TMS system. The first case is the research from Wan Zakaria [28]. Using adaptive force control and a neuro-fuzzy algorithm, he composed a system that can track slight head movement and improve safety. In the same year, L. Ritcher *et al.* [22] suggested a method for controlling the force between the coil and the subject's head by moving the coil in the direction of the end-effector of the robot manipulator or the opposite direction. Prakarn Jaroonsorn *et al.* [31] used the hybrid position/force control to maintain the constant force between the head and the coil while compensating for position error. A. Noccaro *et al.* [21] used impedance control for safe interaction between coil and head.

This study used the hybrid position/force control and the proportional torque control considering the TMS coil's shape. The coil we used in this study has an incurved shape, and we will utilize that characteristic. Because the coil is curved and the human's head is round, better adherence can be made between them. In other words, the facial contact can be utilized between them. It causes facial friction, so it can physically prohibit slight head movement. And because the facial contact makes the distance between the coil and the head smaller, the magnetic stimulation is less affected by the slight head motion.

This research was financially supported by the Korea Institute of Industrial Technology (KITECH) through the In-House Research Program (Development of Core Technologies for a Working Partner Robot in the Manufacturing Field, Grant Number: EO210004).

Jaewoo Kim is with the KITECH school, University of Science and Technology, 34113, Daejeon, Korea and also with the AI·Robotics R&D Department, Korea Institute of Industrial Technology, 15588, Ansan, Korea (beinlove0218@kitech.re.kr; beinlove0218@naver.com).

*Gi-Hun Yang is with the KITECH school, University of Science and Technology, 34113, Daejeon, Korea and also with the AI·Robotics R&D Department, Korea Institute of Industrial Technology, 15588, Ansan, Korea (yanggh@kitech.re.kr).

However, the facial friction also makes it hard for the coil to compensate for the error between its current position and the target treatment point with the small force. That problem was solved by scheduling the magnitude of the force.

This paper is organized as follows: Section II describes the method of calibrating the coordinate system between the camera and the base of the robot manipulator. Section III explains how the trajectory is planned from the initial pose to the target pose until the TMS coil contacts the head. Section IV describes the force and torque control strategy for better accuracy and contact. Section V describes how the robot manipulator tracks the target point when the subject's head is moved. Section VI describes the experimental setup and processes, and the experiment's result is described in section VII. We will discuss our research in section VII. Section VIII summarizes our research and presents possible future works.

## II. THE CALIBRATION BETWEEN THE ROBOT MANIPULATOR AND THE NEURONAVIGATION SYSTEM

To control the robot manipulator by using the coordinate data from the navigation system, first, we should express the poses, which are measured by the camera by the coordinate system of the base of the robot manipulator. The method used for that is explained in this section.

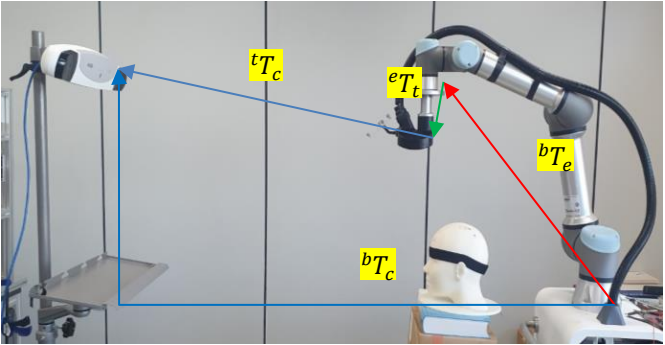

Fig. 1. Transformation matrices of the coordinate. Each matrix describes the coordinate of the arrival point of the arrow by the coordinate system of the starting point of the arrow.

Before explaining the calibration method, the notation of the transformation matrices is described as follows: $^{x}T_{y}$ expresses the pose of $x$ described by the coordinate system of $y$. More clearly, $^{x}T_{y}$ is defined as

$$^{x}T_{y} = \begin{bmatrix} ^{x}R_{y} & ^{x}\mathbf{t}_{y} \\ \mathbf{0} & 1 \end{bmatrix} \quad (1)$$

where $^{x}R_{y}$ is the rotation matrix that describes the orientation of $x$ based on the coordinate system of $y$ and $^{x}\mathbf{t}_{y}$ is the position vector of $x$ based on the coordinate system of $y$. The letters in the location of $x$ and $y$ are as follows: $b$ is the base of the robot manipulator, $c$ is the camera of the navigation system, and $t$ is the tool, that is, the TMS coil. From now on, the location of the TMS coil means the center of the coil's floor, and the coordinate of the TMS coil is the coordinate whose axes are parallel to those of the end-effector of the robot manipulator and whose origin is the location of TMS coil. Using those expressions, we can describe the following equation

$$^{b}T_{c} = {}^{b}T_{e}{}^{e}T_{t}{}^{t}T_{c}. \quad (2)$$

If the robot manipulator and the camera didn't move, theoretically, $^{b}T_{c}$ is constant, irrelevant to the location of the TMS coil. Meanwhile, $^{b}T_{c}$ can be obtained by three values according to (1): The pose of the end-effector of the robot manipulator expressed by the coordinate system of the base of the robot manipulator, the pose of the TMS coil expressed by the coordinate system of the navigation camera, and the CAD information of the TMS coil. Meanwhile, equation (2) can be manipulated as follows:

$$^{b}T_{c} = {}^{b}T_{t}{}^{t}T_{c}. \quad (3)$$

That means,

$$^{b}T_{t} = {}^{b}T_{c}{}^{c}T_{t}. \quad (4)$$

Using (4) after fixing $^{b}T_{c}$, the coordinates estimated by the camera can be converted to the base coordinate system of the robot manipulator.

## III. TRAJECTORY PLANNING

The trajectory was planned to have four phases for making the TMS coil arrive at the target position while avoiding the collision between the subject's head and the robot manipulator. All of the trajectories in this paper are planned using the trapezoidal velocity profile method[26] without the third phase.

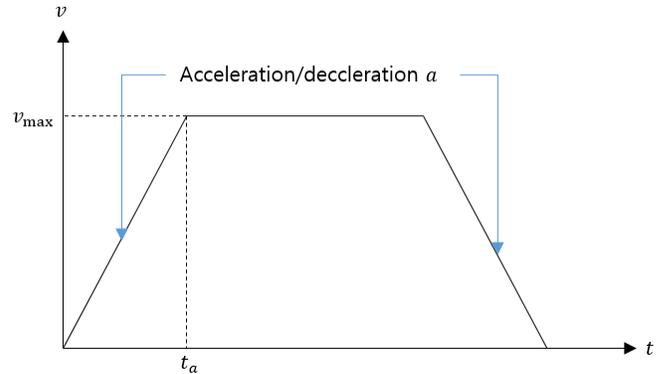

Fig. 2. The velocity-time graph for the trapezoidal velocity profile method

When that method is used for position planning, the maximum velocity $v_{\max}$ and acceleration $a$ are fixed as the graph Fig. 2. Let $s$ be the distance between the initial and final points. Define $t_a$ as the time of the accelerating(decelerating) interval, and $t_f$ as the time of the overall interval of the path. If

$$\frac{v_{\max}^2}{a} > s, \quad (5)$$

that means if the smallest calculated moving distance from the trapezoidal velocity profile with velocity $v_{\max}$ is replaced with $\sqrt{sa}$. The orientation trajectory planning of the first and second phases is similar to the above.

### A. First phase: Approach from the initial position to the spherical surface

Before the robot manipulator starts working, a virtual spherical surface is generated. Its center is 25 mm above the subject's head center, as estimated by the neuronavigation system. Then, the robot manipulator is prohibited from approaching the inside of the sphere to avoid collisions between the robot manipulator and the subject. The reason why the center of the sphere is slightly above the head will be described in Section III. C. Meanwhile, define $\mathbf{x}_i$ as the initial location of the TMS coil, $\mathbf{x}_o$ as the center of the sphere, and $\mathbf{x}_{si}$ the intersection between the spherical surface and the segment $\overline{\mathbf{x}_o \mathbf{x}_i}$ as Fig. 3(a). Let the coil move its position from the initial position to $\mathbf{x}_i$. And let the orientation of the TMS coil be equal to the orientation of the vector $\mathbf{x}_o - \mathbf{x}_i$.

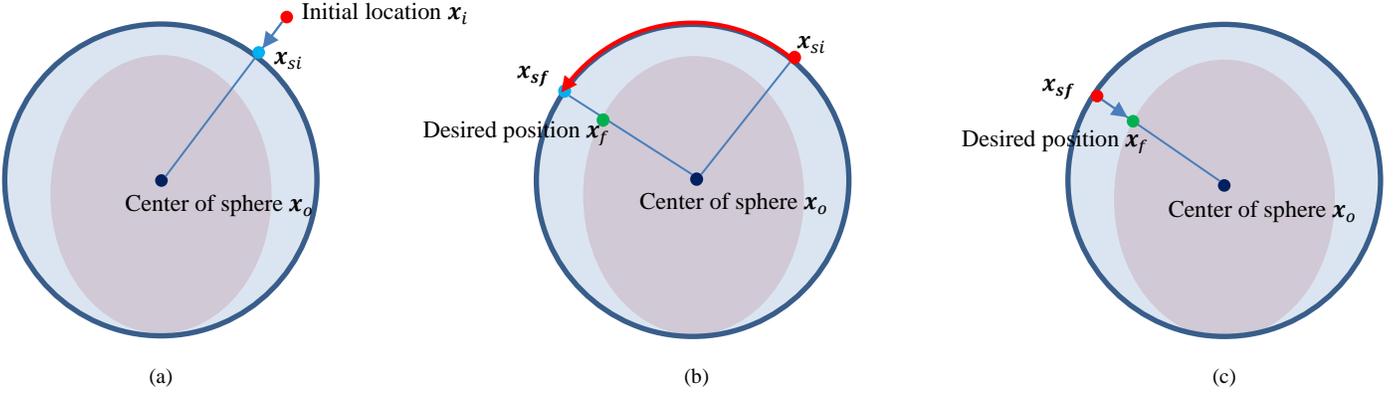

Fig. 3. Trajectories for three phases (a) First phase (b) Second phase (c) Third phase

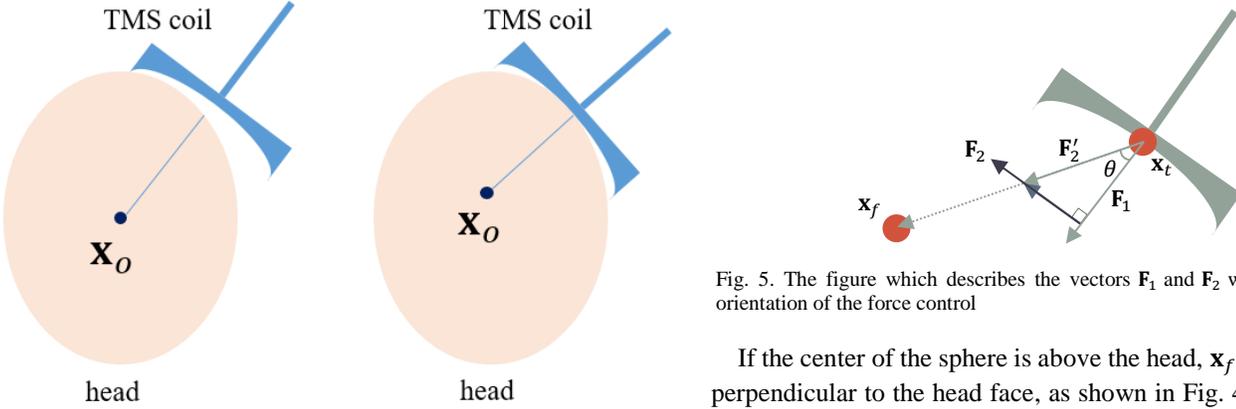

Fig. 4. The last coil pose of the third phase of the sphere's center $\mathbf{x}_o$ is on and the above of the head's center

Fig. 5. The figure which describes the vectors $\mathbf{F}_1$ and $\mathbf{F}_2$ which affects the orientation of the force control

*B. Second phase: The trajectory on the sphere surface*

Let us define $\mathbf{x}_{sf}$ as the intersection between the ray from $\mathbf{x}_o$ to the desired point $\mathbf{x}_f$ and the spherical surface. Because the shortest path between two points on the spherical surface is the shorter arc connecting those[27], the trajectory between $\mathbf{x}_{si}$ and $\mathbf{x}_{sf}$ is decided as the shortest arc of the great circle of the virtual sphere between those two points, as shown in Fig. 3(b). Also, while the coil moves on the spherical surface, the coil faces the center of the sphere.

*C. Third phase: Approaching the subject's head from the point of the spherical surface*

The coil moves from $\mathbf{x}_{sf}$ to $\mathbf{x}_f$ with the same velocity. The orientation of the coil is maintained in this phase. Because the last pose of the next phase is the same as that of this phase, $\mathbf{x}_f - \mathbf{x}_o$ is the orientation before starting the force/torque control. The trajectory explained is expressed graphically in Fig. 3(c).

If the center of the sphere is above the head, $\mathbf{x}_f - \mathbf{x}_o$ is more perpendicular to the head face, as shown in Fig. 4. It helps the coil's center approach near the head surface. The trajectory planning is finished if the coil contacts the subject's head. The system judges whether the coil is contacted by the measured force from the force sensor embedded in the robot manipulator's wrist. If the measured force exceeds the threshold value, the control system recognizes that the coil is in contact with the subject's head.

*D. Fourth phase: The zero adjustment of the force/torque control*

To remove the bias of the force/torque sensor, the sensor is initialized after moving the coil 5 mm from the contact point of the third phase to the third phase in the opposite direction of the coil. After that, the coil is moved to the contact point again to be ready to start the force/torque control.

IV. FORCE AND TORQUE CONTROL FOR ADHERING BETWEEN THE TMS COIL AND HEAD

If the TMS coil is controlled only by the position control, the coil may not be adhered to the subject's head. In this case, the position of the brain aimed by the TMS can vary significantly from the head's movement, even if small. Also, it is not possible to compensate for a relatively small error without considering the interaction between the coil and the head. In this paper, the control strategy is designed to make the coil attach to the head while reducing the error between the target and current coil position by using compliance control, hybrid position-force control, and proportional torque control.

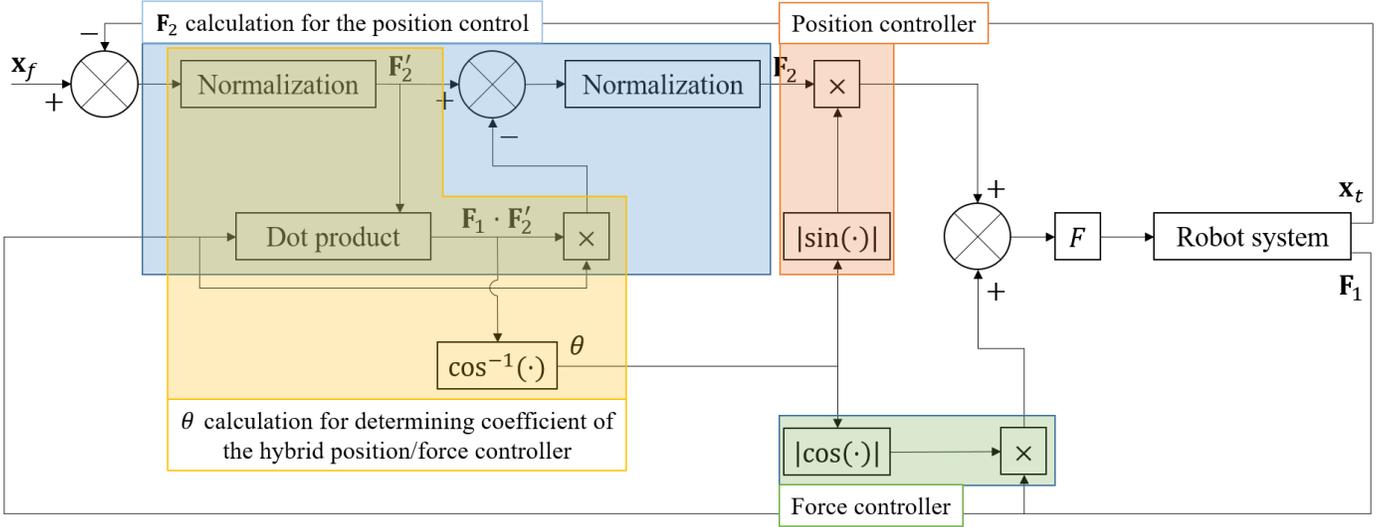

Fig. 6. The block diagram of the force control

## A. Force Control

$x_t$ represents the location of the TMS coil and unit vector $F_1$ represents the orientation in which the coil is oriented. Defined $F_2' \coloneqq (x_t - x_f)/\|x_t - x_f\|$ . For the hybrid position-force control, the unit vector $F_2$ is defined as perpendicular to $F_2'$ and $F_1$ as following:

$$F_2 = \frac{F_2' - (F_1 \cdot F_2')F_1}{\|F_2' - (F_1 \cdot F_2')F_1\|} \quad (6)$$

Meanwhile, suppose that $\theta$ the angle between $F_1$ and $F_2'$. The force that the coil generates is as follows:

$$F = F(F_1|\cos\theta| + F_2|\sin\theta|) \quad (7)$$

where $F$ is the magnitude of the force the coil generates. In this control scheme, the role of $F_1$ is increasing coherence between coil and head, and the role of $F_2$ is compensating error between the current position and the target position. Even if $F_2$ doesn't consider the component of the error, which is parallel to the orientation of the coil, it doesn't matter because $F_1$ will compensate for that by adjusting the coefficients $|\cos\theta|$ and $|\sin\theta|$ from (7). If the target point estimated by the navigation system is located above the head, the coefficient $|\cos\theta|$ may make the error larger because while the force control is trying to reduce the magnitude of the error, which is parallel to $F_1$, that component parallel to $F_2$ may be larger because $F_2$ may be opposite to the orientation of that. In other words, $\theta$ can be near 180°. However, rather than compensating for the error, adhering the coil to the head is more important if the coil is not too far from the estimated target point. The force control explained above is described as a block diagram in Fig. 6.

## B. Torque control

The floor shape of the coil used in this study is incurved and symmetric, as shown in Fig. 7. Therefore, it can widen the contact area between the subject's head and the TMS coil by locating the contact area closer to the center of the coil. By making the contact area larger, the coil can contact stably and resist slight movement of the head better by increasing surface friction. For making the contacting surface near the coil's center, proportional torque control is applied as in the following description.

Suppose the force between the subject's head and the coil's floor is consistent, and the force's distribution on the contact area is uniform. Then, the more the torque acting on the center of the coil's floor is near 0, the more the centroid of the contact area is near the center of the coil's floor. Considering that

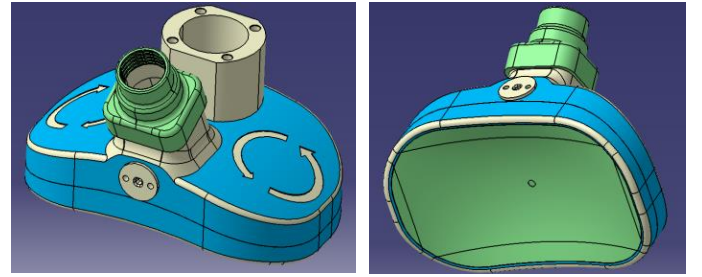

Fig. 7. The shape of the coil

reason, the torque is commanded to the controller as follows:

$$\tau(\tau_c) = -k_p[\tau_{cx}, \tau_{cy}, 0] \quad (8)$$

where $\tau_c = [\tau_{cx}, \tau_{cy}, \tau_{cz}]$ is the measured torque acting on the center of the coil's floor. The coordinate system of $\tau$ and $\tau_c$ are the same as that of the tool. The reason that the $z$ value of $\tau$ is zero is that it can't tilt the coil, so it rarely affects the adhesion.

## C. Compliance control

For the stable interaction between the TMS coil and the head, compliance control is implemented using the URScript function provided by the manufacturer[32]. Force gain is set to the default value, 1.0, and the damping parameter is set to 0.1 in this study. Because the amount of torque generated along the z-axis of the tool coordinate is minimal, compliance control is applied for all orientations except the z-rotation axis of the tool coordinate.

## V. TRACKING THE TARGET POINT WITH TMS COIL

The subject's head may move during the TMS session. If the head movement is not slight, then the target point of the TMS coil has to be changed. If the head doesn't move out fast from the coil, then the coil adheres to the head almost continuously while the head is moving. In this situation, we can track the moved target point by changing the desired point $\mathbf{x}_f$ in the force controller. We showed that it is effective by the experimental result.

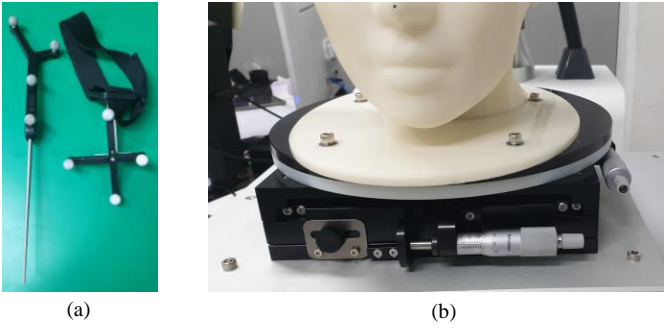

(a)                (b)

Fig. 8. The stylus, the marker-attached headband and the dummy stage

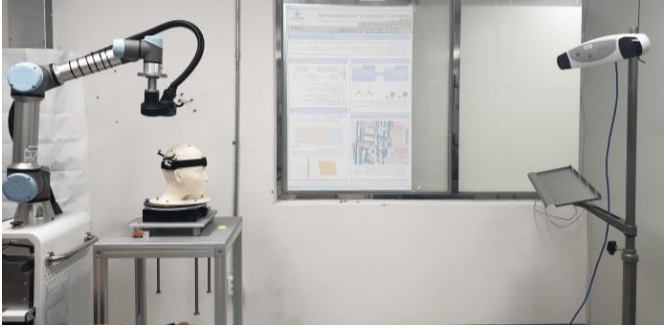

Fig. 9. Experimental environment

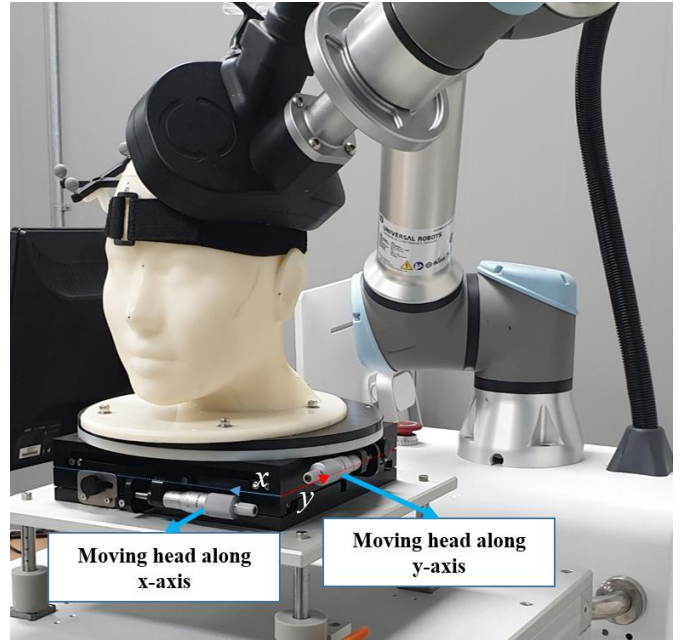

Fig. 10. The coil adhered to the left-dorso lateral of the dummy

## VI. EXPERIMENT

### A. Experimental Setup

Polaris VEGA ST for the camera of our navigation system, the NDI passive sphere for the markers, and the stylus from NDI were used(Fig. 8). One marker was attached to the coil for the calibration between the camera and the robot manipulator. The other one was attached to the headband for estimating the center of the dummy's head and the TMS coil's target point. The manipulator which was used in this research is UR5e from Universal Robot. The robot control program was programmed by Visual Studio C++ 2019 with the ur_rtde library, developed by SDU Robotics. The force/torque sensor embedded in the robot manipulator's wrist measures forces and torques. Because it doesn't have high accuracy or precision [30], it can be examined that the control strategy in this study can be used with inexpensive force/torque sensors. And the TMS coil, which AT&C developed, was used. The distance between the robot manipulator and the camera was around 2.3 ~ 2.5 m, and the dummy is located considering the measurement volume of Polaris VEGA ST[29] and the manipulability of the robot manipulator. The dummy was moved by adjusting the micrometer of the dummy stage(Fig. 8) to test the target point tracking algorithm.

### B. Experimental Protocol

First, the camera estimates the location of the TMS coil using the marker's information. The robot controller read the TMS location information from the camera and calculates the matrix $^bT_c$. The navigation system used the location information of both ears, nose, and middle of the forehead to estimate the location of the head's center and the TMS coil's target point. In this study, we set the target point as the left dorsolateral prefrontal cortex.

After the preparation for the robotized TMS task, the robot manipulator was commanded to locate the TMS coil at the target point. During all of the force control, admittance control was applied, provided by the built-in function of the URScript Programing Language [32]. The force gain of 1.0 and the damping parameter of 0.1 were used for all the experiments. First, it was shown that the position controller of the hybrid position/force controller compensates for the error between the current TMS coil position and the target TMS coil point by comparing two controllers with the same forces: the hybrid position/force controller, which is the same as (7) what we suggested, and the force controller without the position controller, which is $\mathbf{F} = F\mathbf{F}_1$. Second, we applied the suggested force controller with desired forces 5 N, 10 N, 20 N, 30 N, and 40 N without torque control to explore the relation between the magnitude of the desired force and the error of the TMS coil to the desired point and decide how to set the desired force. After choosing, the force controller was applied with the proportional torque controller with the gain $k_p = 0.0, 1.0, 2.0, 4.0,$ and $4.5$ for optimizing the gain. Until parameter $F$ and $k_p$ was chosen, the dummy was fixed during the experiment. After deciding all control parameters for the next experiment, we moved the dummy 7 mm to the $x$ and $y$ axes drawn in Fig. 10 after the coil arrived at the target point. We estimated the new target point from the navigation system and made the robot-manipulated coil to track that. From all experiments, we got the error data between the current and estimated target positions of the TMS coil, the torque applied to the coil's center, and the force applied to the coil.

## VII. RESULT

For convenience, some symbols that are used from this section are defined as follows: $e$ means the positional error magnitude between $\mathbf{x}_t$ and $\mathbf{x}_f$. In other words,

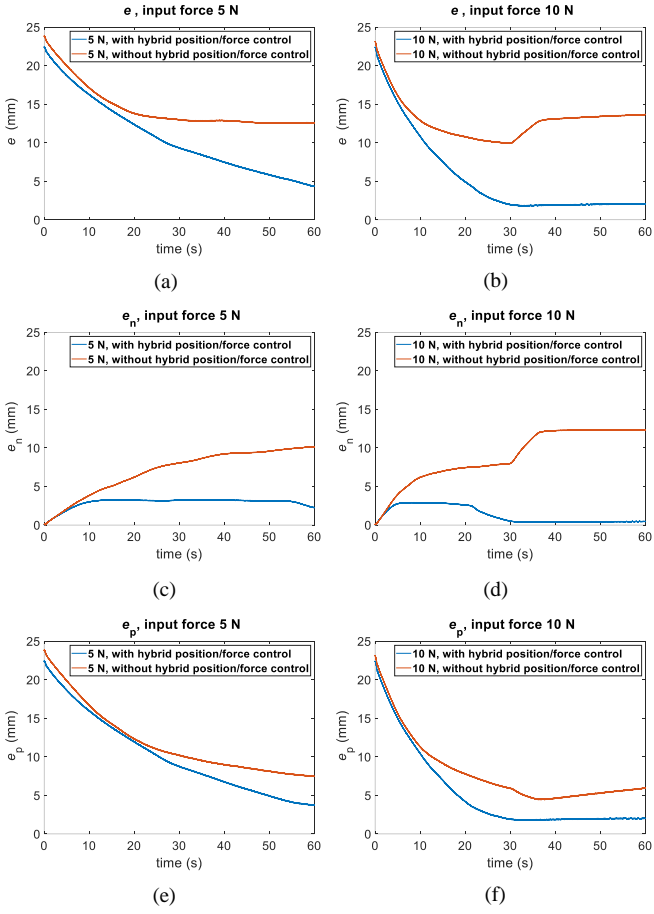

Fig. 11. Comparing e, $e_n$, and $e_p$ between the cases the hybrid/position force is used or not

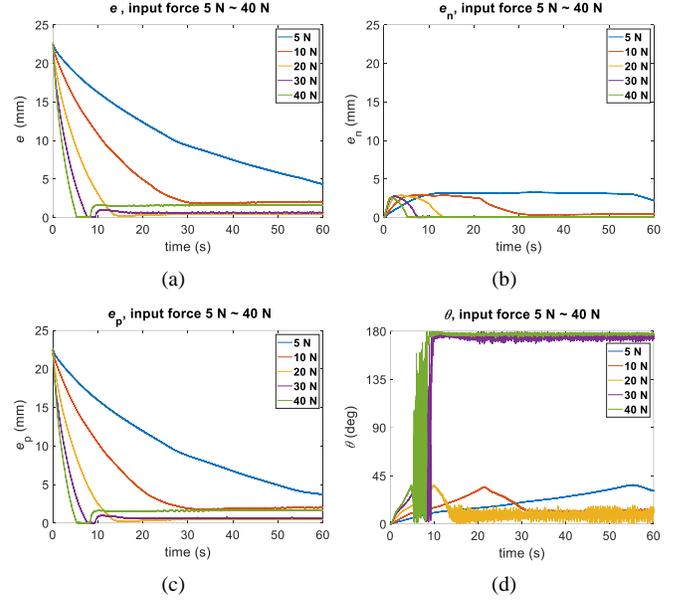

Fig. 12. Comparing $e$, $e_n$, $e_p$, and $\theta$ for each desired forces

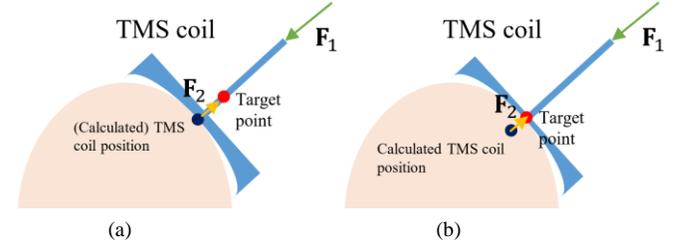

Fig. 13. The two cases which makes $\mathbf{F}_2$ opposite to $\mathbf{F}_1$. (a) The case that the navigation-estimated target point is above the head (b) The case that the calculated TMS coil position is inside the head

$$e = \|\mathbf{x}_f - \mathbf{x}_t\|. \tag{9}$$

$e_n$ and $e_p$ are the magnitude of the error component, which is normal or perpendicular to the TMS coil's orientation. That is,

$$e_n = (\mathbf{x}_f - \mathbf{x}_t) \cdot \mathbf{F}_1 \tag{10}$$

and

$$e_p = (\mathbf{x}_f - \mathbf{x}_t) \cdot \mathbf{F}_2. \tag{11}$$

*A. Error Compensation*

While experimenting with the suggested hybrid position/force controller and without the position controller component, the data of $e$, $e_n$ and $e_p$ were collected during the force control after path planning, as shown in Fig. 11. Torque control was not applied, and the dummy was fixed during the experiment. From Fig. 11(a)-(b), $e$ was smaller with the hybrid position/force controller than without the position controller. Especially from Fig. 11(c)-(f), the hybrid position/force controller compensated $e_n$ more than $e_p$ because of the position controller component of that. The force control component was considered to be an adhesion between the TMS coil and the head instead of compensating for error parallel to $\mathbf{F}_1$.

*B. Setting the desired force of the force controller*

The error data is collected when the different forces were applied to the force controller and described in Fig. 12(a)-(c). There was no torque control and dummy head movement in this experiment. Fig. 12(a) is similar to Fig. 12(c) because, during the experiment, $e_n$ is much smaller than $e_p$. There are two reasons: The value of $e_n$ is theoretically zero when the force control is initialized because when the trajectory planning was ended, the coil was on the segment $\overline{\mathbf{x}_{sf}\mathbf{x}_f}$, and its orientation was the same as $\mathbf{x}_{sf} - \mathbf{x}_f$ from Fig. 3(c). And $e_n$ kept compensated by the position controller component of our suggested force controller. The reason is one of the following: the target point of the TMS coil was above the head because of the navigation system error or the scalp deformation, or the calculated coil's position was inside the head because of the deformation of the head. Each case is described in Fig. 15. If $e_n$ is so small with those cases, $\theta$ can be near 180°. It can be checked by Fig. 12(c)-(d) that if $e_p$ was increased during the experiment, then $\theta$ is near 180°. Because there was almost no probability that the case that can make $\theta$ near 180° during the force control, the reasons that are mentioned are proven.

Meanwhile, from Fig. 12(a), $e$ was reduced faster during the experiment's initial when the desired force was larger while continuously applying large force to the head makes the subject uncomfortable. However, because of the large adhesion surface

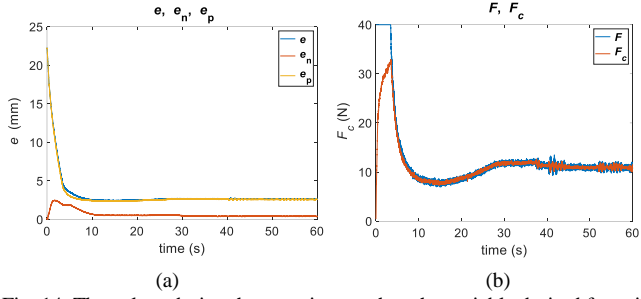

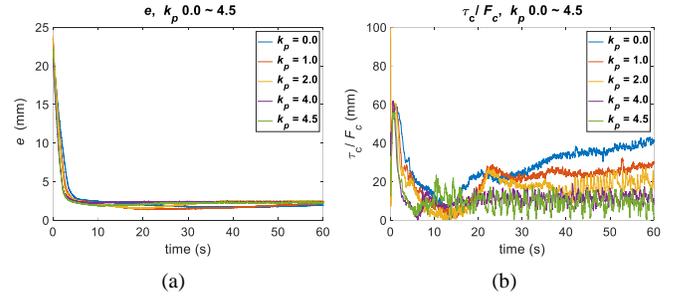

Fig. 14. The values during the experiment when the variable desired force is applied. (a) $e$, $e_n$, and $e_p$ (b) $F$ and $F_c$

Fig. 16. The experimental result for each $k_p$ (a) $e$ (b) $\tau_c/F_c$

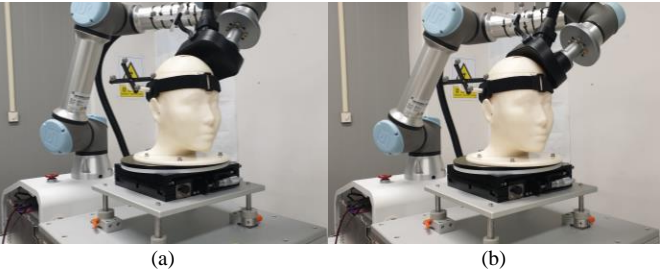

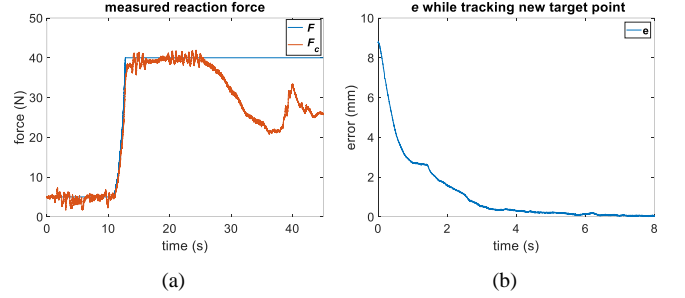

Fig. 15. The TMS coil and the head (a) When the force control is started (b) When the coil adhered to the head

Fig. 17. The results of the tracking experiment (a) Measured reaction force between the TMS coil and the dummy head while the head is moving (b) The error between the current coil position and new target position after moving the dummy

between the coil and the head from the shape of the incurved floor of the coil, it is very slow to use low force for the force control to compensate $e$. For that reason, the desired force is set like the following equation for faster error reduction and selecting a suitable force for interacting with the head and the coil:

$$F = F(e) = \begin{cases} 40 & \text{if } e > 0.2e_o \\ 350(e/e_o) - 30 & \text{if } 0.1e_o < e < 0.2e_o \\ 5 & \text{if } e \leq 0.1e_o \end{cases} \quad (12)$$

where $e_o$ is the initial value of the $e$ after starting the force control. If the $e$ value was larger than 20% of $e_o$, $F$ was fixed to 40 N to reduce e fast. If the $e$ value was between 10 % and 20% of $e_o$, $F$ was reduced linearly by $e$ for reducing the interacting force between the coil and the head. The minimum value of $F$ was set to 5 N to maintain the coil and head cohesion. Fig. 16 shows the $e$, $e_n$, $e_p$ and the interaction force $F_c$ between the head and the coil, measured by the force/torque sensor embedded in the robot manipulator, during the experiment when the desired force was following (12).

From Fig. 14(b), the desired force $F$ was much larger than $F_c$ during the initial term of the experiment, because there were many vacancies between the coil and the head when the force control was started as Fig. 15(a), much of the force commanded for the force controller was used for moving the robot manipulator rather than interacting between the coil and the head.

From Fig. 14, the error was reduced fast as intended, and the converged $e$ value was under 3 mm. The large force over 20 N is applied to the head briefly, less than 10 seconds. From those results, (12) is used for the desired force of our force controller.

### C. Setting $k_p$ for the proportional torque control

After deciding on $F$, the same experiments were done with the force and torque control and got the values $e$ and $\tau_c/F_c$ for each $k_p$ during the force control. The values are described in Fig. 16. Because the applied torque on the object was proportional to the applied force on it, we can reject the effect of the control force from $\tau_c$ by dividing it by $F_c$. So the value $\tau_c/F_c$ can be compared by the cases that applied different forces to check how close the contact area between the dummy and the coil was to the center of the coil.

From Fig. 16(a), it can be seen that larger $k_p$ reduces $e$ slower at the first time of the experiment and makes the steady-state $e$ value larger. However, it is not very meaningful: The difference between the maximum and minimum time to the first time the value of e decreases below 5 mm is 1.2440 seconds. The maximum error difference between each $k_p$ after 10 seconds from the start of the experiment is less than 1.1 mm. So, $k_p$ was decided by the $\tau_c/F_c$ values, as shown in Fig. 16(b). It is shown that when $k_p = 4.0$, lower $\tau_c/F_c$ value was obtained than the same value of the less $k_p$ after 20 seconds from the initial of the experiment. Meanwhile, using $k_p$ Larger than 4.0 made a large oscillation to that value, making the coil pose unstable. From those results, $k_p = 4.0$ is chosen.

### D. Tracking the moved target point

In this experiment section, the head was moved slowly while the coil was adhering to the head. During the experiment, the measured reaction force $F_c$ was collected between the TMS coil and the head and the desired force $F$; those data are described in Fig. 17(a). While moving, $F_c$ didn't lower abruptly. That means the coil continuously interacts with the head. From around 12 seconds of the graph of Fig. 17(a), $F$ increased suddenly because of the variance of $e$. After that, $F_c$ didn't follow $F$ well because much of the applied force was used for moving for compensating the error. However, it doesn't cause problems because error information before getting a new target point is meaningless after moving the head and interacting between the coil.

After the head movement, the coil tracked the moved target point with the force control. Because the adhesion between the coil and the head was maintained during the movement, tracking the target point with the suggested force/torque controller was possible. The error information is described in Fig. 17(b).

## VIII. DISCUSSIONS

The adhesion between the subject's head and the TMS coil is improved by using the characteristics of the coil's floor, which can help reduce the effect of small movements of the subject's head on the robotized TMS task. To be ready to realize that goal, we did the calibration work and trajectory planning. The transformation matrix from the camera coordinate system to the robot manipulator's base coordinate system is calculated from the calibration work. With that matrix, coordinate data from the navigation system can be expressed by the robot manipulator's base coordinate system. While generating the trajectory, the sphere, including the subject's head, is generated to prevent collision between the head and the coil by preventing the coil from invading the sphere. After following the planned trajectory, several force/torque control strategies are applied for better adherence: Compliance control for the interaction between coil and head and proportional torque control to make the center of contact area between the head and coil near the center of the coil's floor. While using those strategies, the magnitude of the force used for those is scheduled by $e$ for reducing $e$ faster. By that scheduling, the low $e$ can be maintained with the lower force after applying large force for a short time.

In the system used in our experiment, the force/torque sensor embedded inside the UR5e manipulator is used, with no additional external sensors. If there are external forces, for example, the tension from the TMS coil's cable, that affects the force measuring enormously from the embedded force/torque sensor, additional external sensors may be needed.

For future work, real-time tracking will be applied to the present system with the control strategy developed in this study. By embodying that function, subjects of our system may feel less pressured to fix their head.

## IX. CONCLUSION

The force/torque controller for the robotized TMS system using the incurved coil is presented in this study. Hybrid position-force control is used to make better cohesion between the TMS coil and the subject's head and better accuracy between the current and the desired point of the TMS coil. The proportional torque control is used to reduce the torque applied on the center of the coil, which makes the contact area between the coil and the head inside the coil and makes the contact area wider. It can make the contact more stable.

In this paragraph, the Cartesian coordinate, which is mentioned, is the tool coordinate system. From the experimental results, the larger force applied for the force control of the robot manipulator reduces $e$ faster. However, if a large force is applied to the subjects, it can cause uncomfortability to them. Nevertheless, applying low force for the robot manipulator's force control causes slow compensation of the error between the coil's current and target positions because of the large contact area between the coil and the head. Because of those reasons, the applied force is scheduled by the $e$ value. If $e$ is reduced, the applied force is reduced until 5 N. The minimum force is set for maintaining the adhesion between the coil and the head. The $z$-axis component of the error may be larger during the experiment if the target point estimated from the navigator is above the head. However, it can be sacrificed to ensure firm adhesion between the coil and the head.

Meanwhile, while the head moves during the TMS session, the coil continuously adheres to the head if it moves slowly. Because of that, the coil can track the moved target point using our force control strategy.

9[20] Q. Cheng *et al.*, "Trajectory planning of transcranial magnetic stimulation manipulator based on time-safety collision optimization," *Rob. Auton. Syst.*, vol. 152, p. 104039, Jun. 2022.

[21] A. Noccaro *et al.*, "Development and validation of a novel calibration methodology and Control Approach for robot-aided transcranial magnetic stimulation (TMS)," *IEEE. Trans. Biomed. Eng.*, vol. 68, no. 5, pp. 1589–1600, May. 2021.

[22] L. Richter *et al.*, "Hand-assisted positioning and contact pressure control for motion compensated robotized transcranial magnetic stimulation," *Int. J. Comput. Assist. Radiol. Surg.*, vol. 7, no. 6, pp. 845–852, Mar. 2012.

[23] G. Pennimpede *et al.*, "Hot spot hound: A novel robot-assisted platform for enhancing TMS Performance," *in Conf. Proc. IEEE Eng. Med. Biol. Soc.*, 2013.

[24] L. Richter *et al.*, "Towards direct head navigation for robot-guided transcranial magnetic stimulation using 3D laserscans: Idea, setup and feasibility," *in Conf. Proc. IEEE Eng. Med. Biol. Soc.*, 2010.

[25] R. Ginhoux *et al.*, "A custom robot for transcranial magnetic stimulation: First assessment on healthy subjects," *in Conf. Proc. IEEE Eng. Med. Biol. Soc.*, 2013.

[26] B. Siciliano *et al.*, *Robotics: Modelling, Planning and Control*. London, United Kingdom: Springer-Verlag, 2010, pp. 165-167

[27] J.-H. Eschenburg, *Geometry-Intuition and Concepts: Imaging, understanding, thinking beyond. An introduction for students.* Wiesbaden, Germany: Springer Nature, 2022, pp. 109-110

[28] W. Zakaria, "Force-controlled transcranial magnetic stimulation(TMS) robotic system," Ph. D. dissertation, Dept. Mech. Sys. Eng., Newcastle Univ., Tyne and Wear, England, 2012

[29] NDI. Inc., POLARIS Vega® ST. [Online]. Available: https://www.ndigital.com/optical-measurement-technology/polaris-vega-st/

[30] Universal Robots. Inc., UR5e Product Fact Sheet. [Online]. Available: https://www.universal-robots.com/media/1807465/ur5e-rgb-fact-sheet-landscape-a4.pdf

[31] P. Jaroonsorn *et al.*, "Robot-assisted transcranial magnetic stimulation using hybrid position/force control," *Adv. Rob.*, vol. 32, no. 24, pp. 1559-1570, Dec. 2020.

[32] Universal Robots. Inc., THE URScript Programming Language for e-Series. [Online]. Available: https://s3-eu-west-1.amazonaws.com/ur-support-site/115824/scriptManual_SW5.11.pdf